\documentclass[letterpaper, 10 pt, conference]{ieeeconf}  

\IEEEoverridecommandlockouts                              

\usepackage{amssymb}
\usepackage{bbm}
\usepackage{graphicx}
\usepackage{amsmath}
\usepackage{amssymb}
\usepackage{tabularx}
\usepackage{slashbox}
\usepackage[normalem]{ulem} 
\usepackage{subfigure}

\usepackage[nolist]{acronym}
\usepackage[textsize=small]{todonotes}

\long\def\invis#1{}
\usepackage[normalem]{ulem}

\long\def\greg#1{}

\newcommand\gderror[1]{
  \typeout{--------------------------------------------------------------------}
  \typeout{------- #1 ---------}
  \typeout{--------------------------------------------------------------------}
  {\bf #1}
}
\newcounter{gdTmp}
\newcounter{gdLastCount}
\newcommand\maxpage[2][Error]{  
\ifnum\value{page}>#2
   \gderror{\Large \bf On page {\thepage} we are past page #1 (too long).   #2 }
\else\fi
}
\newcommand\maxpageSinceLast[2][Error]{  
\ifnum \numexpr \value{page} - \value{gdLastCount}\relax>#2
   \gderror{Exceeds max length #2 pages. Page \thepage: #1}
\thepage\else\fi
\setcounter{gdLastCount}{\value{page}}
}

\maxpage[{\color{red}\Large Main text is too long.}]{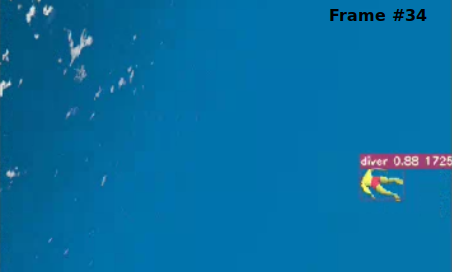}

\usepackage{caption}
\usepackage{hyperref}
\usepackage{amsmath, amsfonts}
\usepackage{dsfont}
\title{\LARGE \bf
	A comparison of RL-based and PID controllers for 6-DOF swimming robots: hybrid underwater object tracking
}

\author{Faraz Lotfi$^{1}$, Khalil Virji$^{1}$, 
Nicholas Dudek$^{2}$, and Gregory Dudek$^{1}$
	\thanks{$^{1}$Mobile Robotics Lab (MRL), School of Computer Science, McGill University,
		Montreal, Canada
		{\tt\small \{f.lotfi, kvirji, dudek\}@cim.mcgill.ca}}
  \thanks{$^{2}$Independent Robotics Inc,
		Montreal, Canada
		{\tt\small \{nick.dudek\}@independentrobotics.com}}%
}

\begin{document}

	\maketitle
	\thispagestyle{empty}
	\pagestyle{empty}

	\begin{abstract}
In this paper, we present an exploration and assessment of employing a centralized deep Q-network (DQN) controller as a substitute for the prevalent use of PID controllers in the context of 6DOF swimming robots. Our primary focus centers on illustrating this transition with the specific case of underwater object tracking. DQN offers advantages such as data efficiency and off-policy learning, while remaining simpler to implement than other reinforcement learning methods. Given the absence of a dynamic model for our robot, we propose an RL agent to control this multi-input-multi-output (MIMO) system, where a centralized controller may offer more robust control than distinct PIDs. Our approach involves initially using classical controllers for safe exploration, then gradually shifting to DQN to take full control of the robot.

We divide the underwater tracking task into vision and control modules. We use established methods for vision-based tracking and introduce a centralized DQN controller. By transmitting bounding box data from the vision module to the control module, we enable adaptation to various objects and effortless vision system replacement. Furthermore, dealing with low-dimensional data facilitates cost-effective online learning for the controller. Our experiments, conducted within a Unity-based simulator, validate the effectiveness of a centralized RL agent over separated PID controllers, showcasing the applicability of our framework for training the underwater RL agent and improved performance compared to traditional control methods. 
The code for both real and simulation implementations is at https://github.com/FARAZLOTFI/underwater-object-tracking. 
 \end{abstract}


	\section{Introduction}



RL-based control is recognized for its flexibility and generality, yet its practical application remains limited in complex underwater robotics, specifically
for vehicles with complex dynamics and
actions spaces. Simple classic controllers, such as those which linearly combine proportional, derivative, and integral error estimators (PID), are still widely used due to their simplicity,
relative understandably and relative robustness. 
In a multiple-input multiple-output (MIMO) system, however, adjusting one variable can affect others, notable in underwater navigation scenarios where linear velocity, yaw rate, and pitch rate must be carefully balanced. To address this challenge, we propose employing a single RL agent to control pitch and yaw rates while retaining a PID controller for linear velocity. This combination leverages the strengths of both approaches and allows the RL agent to adapt to PID controller performance
making both the PID and RL components mutually 
more usable than either alone. Furthermore, formulating the problem as an RL task enables the incorporation of crucial constraints such as torque limits that cannot be effectively handled by separate PID controllers. 

\begin{figure}[]
	\begin{center}
	   \includegraphics[width=0.9\linewidth]{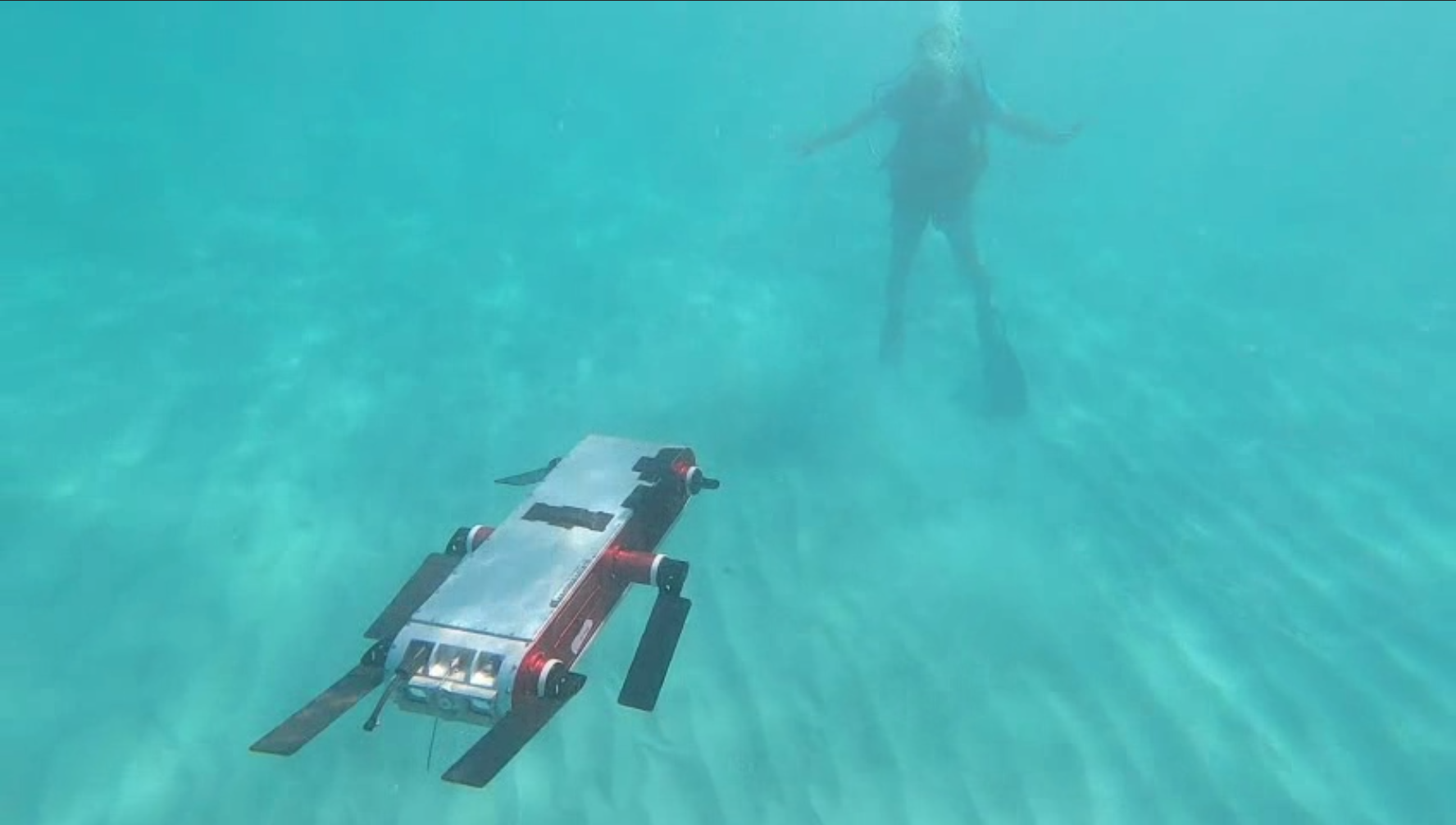}
	\end{center}
	\caption{The robot with six legs for precise control and maneuvering in underwater tasks. }
	\label{aqua_robot}
    \end{figure}

Although RL controllers have been used for previous underwater robots~\cite{el2013two, carlucho2018adaptive}, the 6DOF hexapod robot presents a higher level of complexity due to coordinating six distinct legs for swimming~\cite{aqua,aqua_gait}. Fig.\ref{aqua_robot} shows our six-legged robot called Aqua, which requires precise control\cite{aqua_gait} for executing gaits and tasks involving angular and translational rates. A controller generates these rates and feeds them into the gait controller. Previous underwater swimming robots rely on separate PID controllers for yaw-pitch rates due to the lack of a proper model~\cite{ours, Junaed_generic, sattar2009vision, travis_goal_conditioned, junaed_last}. However, the interactions in the six-legged Aqua states necessitate a centralized PID controller for improved performance, contingent on a dynamical system model~\cite{garrido2012centralized, ma2022position}.  

Considering the task of underwater object tracking, vision-based methods are the most popular approach due to their adaptability and lack of strict assumptions~\cite{ours, lab_pre_research}. To reduce computational costs and enable online updates of the model, we break down the problem into vision and control components~\cite{break_down} rather than using an End-to-End architecture~\cite{DrQ,toromanoff2020end}.
In addition, if the system involves a time-delay, it's important to consider the Markov property more carefully to ensure accurate tracking~\cite{delay_paper1,delay_paper2}.

In this paper, we demonstrate that a centralized DQN controller can outperform the traditional approach of using separated PID controllers. We address the issue of sample complexity~\cite{transfer_learning,buckman2018sample} by presenting a framework that trains the RL agent using existing classic controllers to safely explore the state-action space and provide guidance when the agent's policy is not yet converged. Finally, we benefit from a spiral search-inspired approach~\cite{burlington1999spiral, ours} for target recovery, which is critical in preventing the need to re-run trials frequently after failures. We employ DQN to minimize the burden of real-world implementation while showcasing the practical applications of RL-based controllers.  

To summarize, the primary contributions of this work are:\\
    -An examination of the feasibility of using a centralized RL controller in place of the conventional, segregated PID controllers for a 6DOF swimming robot.\\
    -The implementation of an online improvement loop that can effectively handle distribution shifts, uncertainties, and variations in system parameters. \\
    -The use of a single camera to directly generate pitch and yaw rates, thereby enabling the target 
    view to remain centered without the need for additional state estimation.\\
    -A general framework for optimizing the training of an RL-based controller via the use of an existing classical control approach, simplifying the exploration of the state-action space.
    
The paper is structured as follows: Sec. 2 covers related work, Sec. 3 presents the proposed framework and associated heuristics for enhancing the RL agent's performance, Sec. 4 discusses details for robust performance and includes the studies carried out on the simulator, and Sec. 5 provides concluding remarks.
\section{Related Work}


In this section, we briefly review the existing literature on vision and control applied to
diver tracking~\cite{tracking2020sattar,SattarDudekFourier,NAD2016214}. Our study uses deep learning-based object detection models like YOLO~\cite{redmon2016you}, R-CNN~\cite{girshick2014rich}, and SSD~\cite{liu2016ssd}, which are widely used for real-world tasks and being continually improved~\cite{wang2022yolov7, ren2015faster}. Studies have demonstrated that YOLO is reliable for underwater object detection~\cite{ours, lab_pre_research, de2021video}. To enhance temporal stability during tracking, we incorporate the methodology previous work~\cite{ours,tracking2020sattar}, merging YOLO with SORT. Additionally, we employ an RL-based controller to guide targets towards optimal detection regions, ensuring the proficient performance of our object detector.

The controller component of underwater object tracking is not as well-established as the vision component. Previous research in this field has relied on widely-used PID controllers, as discussed in~\cite{lab_pre_research, ours, junaed_last} where the first two share a similar perception and control approach to our paper. Although there has been active research in other applications on tuning PID controllers based on solving optimization problems~\cite{tuning_PIDs,SHUPRAJHAA2022109450}, these controllers still have limitations in taking into account the variables' interdependence which requires a system model, and they have not yet found their way into use for complex underwater robots. A decentralized approach often leads to conflicts between separate PID subsystems and an increased likelihood of failure, and potentially higher energy consumption. 

Controllers that employ deep reinforcement learning can provide high performance~\cite{singh2022reinforcement} and are shown to deliver better results compared to the traditional PID controllers~\cite{krishna2022epersist}. 
Deep RL has been used for both continuous action space~\cite{lillicrap2015continuous,schulman2017proximal} and the discretized case~\cite{watkins1992q,van2016deep}. Although the former approaches provide high accuracy, the latter ones are more data efficient~\cite{su2017sample}. Moreover, it has been shown that discretized action space methods can deliver comparable results to continuous ones, even for high dimensional spaces~\cite{tang2020discretizing,van2020q}. In this study, we utilize the double Q-learning technique from~\cite{van2016deep} and discretize the robot's action space. Our results demonstrate that a centralized RL agent can outperform traditional PIDs, with potential for further performance enhancement using advanced RL approaches. 

In addition to the issues noted above, the use of RL raises the concern of balancing exploitation and exploration, particularly when safety is a priority. This is a widely questioned issue in the field, and a recent survey is available~\cite{gu2022review}. There are several ways to address safety in RL, including incorporating it into the optimization process~\cite{tamar2012policy,lutjens2019safe, ding2020natural} by defining cost functions or gradients with safety constraints, or leveraging external knowledge such as expert demonstrations~\cite{abbeel2010autonomous, garcia2012safe}. In our approach, we adopt the latter method, but instead of relying on expert demonstrations, we utilize the generated actions to safely explore the state-action space while ensuring system stability.  

    \begin{figure}[]
	\begin{center}
	   \includegraphics[width=0.9\linewidth]{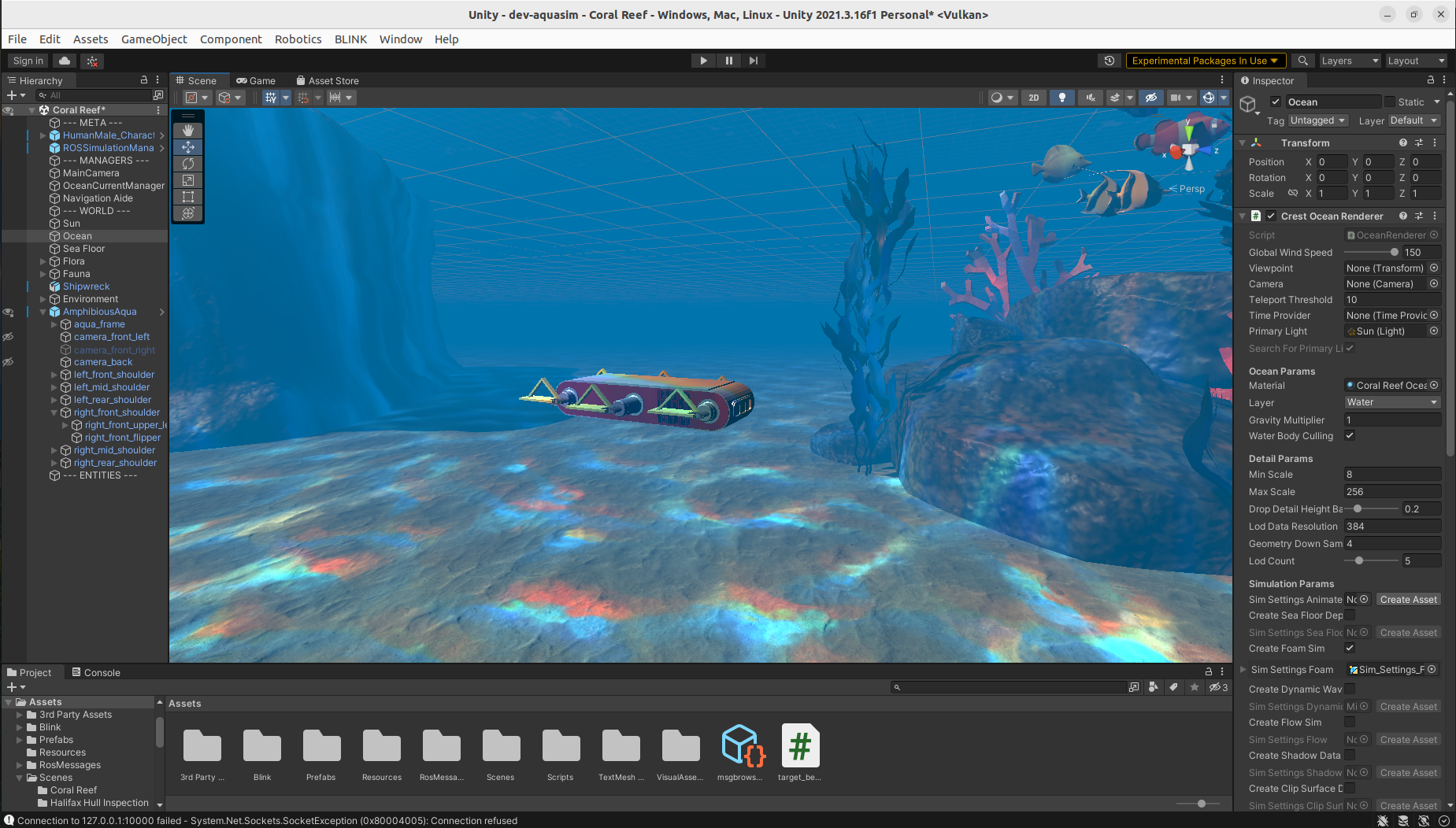}
	\end{center}
	\caption{Marine robotics simulator used for controlled evaluation~\cite{IRsim} (see Sec~\ref{simsec}).}
	\label{fig_sim_1}
    \end{figure}
	\section{Proposed Framework}
\subsection{Vision Module}
 We have adopted the methodology outlined in ~\cite{ours}, using YOLOv7 for object detection, and then employing SORT~\cite{bewley2016simple} for detection matching and implementing tracking-by-detection. SORT is based on the Kalman filter, and estimates states that may not be directly measurable or whose measurements are subject to noise. The system model is assumed to be:
\begin{eqnarray}
    &x_{k|k-1} = F_k x_{k-1|k-1}\\ 
    &y_k = H_k x_{k|k-1}
\end{eqnarray}
where the system state vector is denoted by $x \in \Re^{n}$, while the output vector is represented by $y \in \Re^{m}$. The system model is defined by $F_k$ and $H_k$, which describe the evolution of the system and the relationship between the output and the state, respectively. 
In our vision module, the bounding box coordinates are provided as observations by the object detector, and the system state vector is defined as $[x_{c}, y_{c}, a_{b}, r_{b}, c_{d}, \dot{x}_{c}, \dot{y}_{c}, \dot{a}_{b}, \dot{r}_{b}, \dot{c}_{d}]^{T}$. This vector represents the center point of the bounding box coordinates relative to the image plane center, the area and scale of the bounding box, the detection confidence, and the velocities of all states. It is worth noting that we also filter the detection confidence data, as its measurement is noisy. Additionally, we utilize a linear constant velocity model to carry out the prediction step. Moreover, SORT efficiently handles multi-object tracking using the Intersection over Union (IoU) metric for matching, despite our focus on single-object tracking.
	\subsection{RL Controller}

        As a time-delay system, the 6DOF hexapod robot cannot be effectively addressed using turn-based RL, which violates the Markov property and leads to suboptimal policies or even unstable systems. To address this issue, alternative methods such as real-time RL~\cite{ramstedt2019real} must be explored.
        The interaction between a robot and its environment can be modeled using a Markov Decision Process. In this framework, the probability of transitioning to the next state, $s_{t+1}$, can be calculated as follows:
        \begin{equation}
            \kappa(s_{t+1}|s_t) = \int_A p(s_{s+1}|s_t, a)\pi(a|s_t)da
            \label{turn_based_dist}
        \end{equation}
        Here, $A$ denotes the action space, $s_t$ is the current state, $a_t$ is the current action being taken, $\pi$ refers to the policy being used to generate commands, and $p$ represents the world model. The primary objective is to maximize the expected value of the discounted sum of rewards over time; based on this, our reward function is defined as follows:
        \begin{equation}
        \bar{r}(s_t) = \int_A r(s_t,a) \pi(a|s_t)da 
        \end{equation}
        This model is widely adopted in many benchmark environments of OpenAIGym/MuJoCo~\cite{openai,mujoco}. The interaction is called turn-based as the environment temporarily halts while the agent chooses an action, and the agent awaits a fresh observation from the environment before resuming. In real-time RL, the mentioned equations are re-defined as:
        \begin{eqnarray}
            &\kappa(s_{t+1}, a_{t+1}|s_t, a_t) = p(s_{s+1}|s_t, a_t)\pi(a_{t+1}|s_t, a_t) \\
            &\bar{r}(s_t, a_t) = r(s_t, a_t)
        \end{eqnarray}
        
\begin{figure}[]
        \centering
        \subfigure{\includegraphics[width=0.33\textwidth]{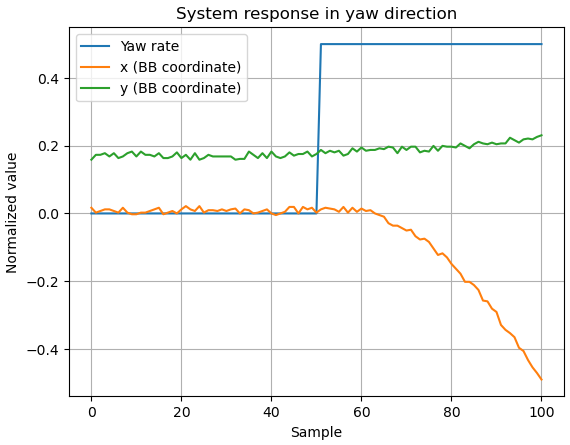}} 
        \subfigure{\includegraphics[width=0.33\textwidth]{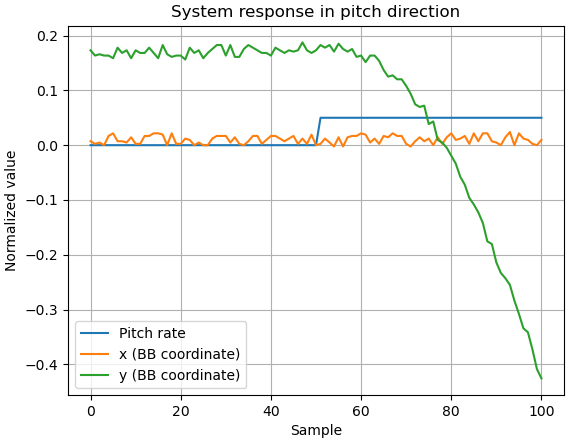}} 
        \caption{This plot demonstrates how the tracking system responds to constant yaw and pitch rates revealing the existing delay. }
        \label{delay}
        \end{figure}

        To account for concurrent environment and robot evolution, the policy relies on input in the form of $(s_t, a_t)$, representing the current system state. However, a response delay violates the Markov property. Fig. \ref{delay} illustrates the robot's behavior under a constant yaw/pitch rate, fixed target, and constant forward velocity, revealing a noticeable lag before action effects. To address this, we augment a history of states and actions to the current state, forming a new Markov state space. This history's length must exceed the delay time. In our experiments, we demonstrate that insufficient history consideration can lead to suboptimal policies.
        
\begin{figure*}[]
        \begin{minipage}[c]{0.5\linewidth}
        \begin{center}
	   \includegraphics[width=1\linewidth]{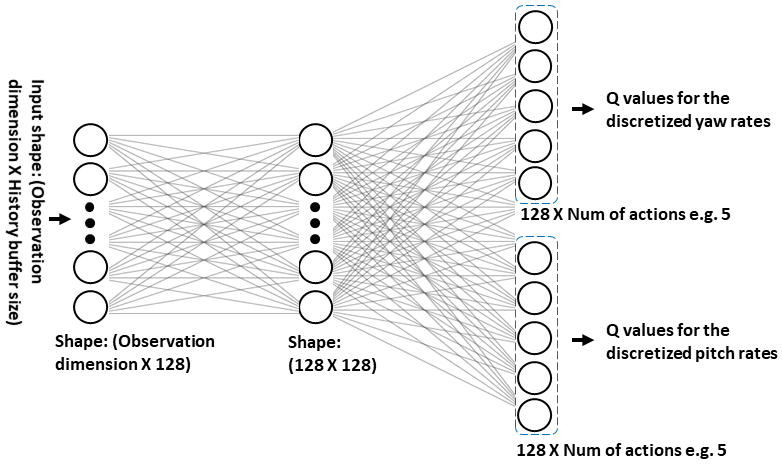}
	\end{center}
	\caption{The network model with two heads generates $Q$ values for pitch and yaw rates.}
	\label{fig_nn}
        \end{minipage}
        \begin{minipage}[c]{0.5\linewidth}
        \centering
        \includegraphics[width=1\textwidth]{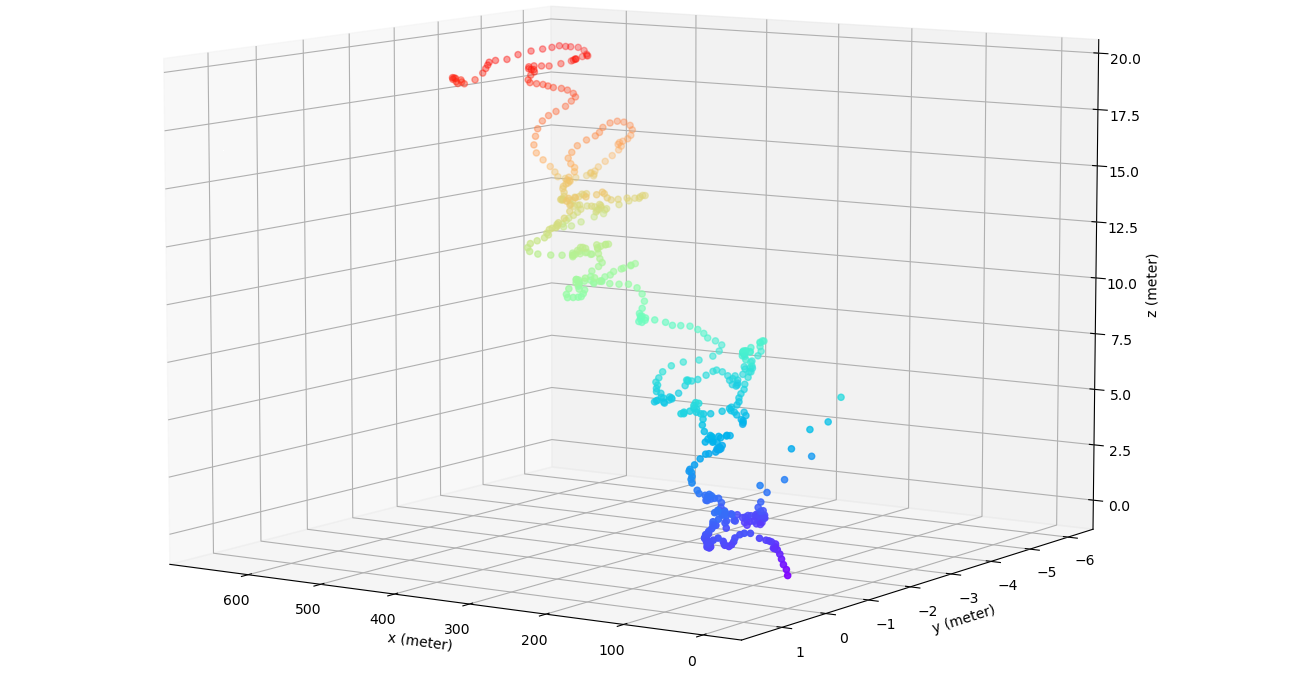}
        \caption{The graph displays the robot's 3D trajectory in the initial study's diver tracking task, intentionally incorporating the diver's random movements to enhance the task's complexity.}
        \label{diver_behavior}
        \end{minipage}
        \end{figure*}

        Fig. \ref{fig_nn} illustrates the network model utilized for generating control commands. As reported in~\cite{travis_goal_conditioned}, a similar architecture was proposed while following an imitation learning scheme. Unlike typical deep Q networks, our model has separate heads for generating yaw and pitch rates. This approach results in two distinct sets of Q values for the two directions, rather than outputting all possible combinations of actions. By reducing the number of outputs for higher dimensions of the action space, we avoid the issue of an exploding number of outputs. We believe that sharing parameters and using a common reward function are crucial for realizing a centralized controller.
        The model takes a state vector as input, which contains $[x_{t-1}, y_{t-1}, area_{t-1}, {v_l}_{t-1}, a_{t-1}]$. These values represent the center point and area of the bounding box, the linear velocity command generated by a PID controller, and the previous action vector containing yaw and pitch rates. Note, the filtering technique discussed in the previous subsection provides the first three state values. 
        
        
        In the context of our RL-based diver tracking problem, it is essential to design a suitable reward function. To this end, we propose a shared reward function for both heads in our architecture. Our reward function, as defined in equation (\ref{eq_reward}), aims to guide the robot in keeping the target centered on the image plane. It is important to note that our approach to this problem utilizes non-episodic/continual RL.
        \begin{equation}
        r(t) = \frac{ \mu}{\lvert x_{c}(t)\lvert + \mu} + \frac{ \lambda}{\lvert y_{c}(t)\lvert + \lambda}       
        \label{eq_reward}
        \end{equation} 
        where $\mu$ and $\lambda$ are used to control the sharpness of the reward function. Using the double DQN approach, we then update the $Q$ values by applying this equation:
        \begin{eqnarray} 
        &\resizebox{.85\hsize}{!}{$Q_C (s_t,a_t) = r_t + \gamma Q_{T}(s_{t+1}, argmax_{a_{t+1}}Q_C(s_{t+1},a_{t+1}))$} \\ 
        &\theta_T = \tau \theta_C + (1 - \tau)\theta_T       
        \end{eqnarray}
        where $Q_C$ and $Q_T$ are the $Q$ values obtained from the current and target networks, respectively. $r_t$ represents the immediate reward obtained at time step $t$, while $s_t$ and $a_t$ correspond to the current state and action taken. The weights of the current and target networks are denoted by $\theta_C$ and $\theta_T$, respectively.
        The remaining parameters are reported in Table \ref{parameters}.   


        \begin{table}[]
        \small
        \centering
        \begin{tabularx}{0.47\textwidth}{| c | c | c | c | c | c | c | }
        \hline
        \textbf{\textbf{Param}} & $\eta$ & $\gamma$ & $\tau$ & \scriptsize{Batch size} & \scriptsize{ERM size} & $\mu$, $\lambda$   \\
         \hline
         \textbf{Value} & 1e-5 & 0.99 & 0.001 & 50 & 2000 & 0.1  \\ 
         \hline
         \end{tabularx}   
         \caption{Parameters used in experiments}
         \label{parameters}
        \end{table}
        
        Furthermore, the simulator utilized in this study lacks the capability to run multiple agents and environments simultaneously, resulting in a notable extension of the training duration. However, it's worth noting that such challenges are typical in practical RL agent training, where factors like instability and safety, as highlighted in a recent review~\cite{gu2022review}, must also be addressed. This limitation actually contributes to the simulator's realism, mirroring the challenges encountered in real-world RL training scenarios.
        
        To overcome these challenges, we introduce a systematic training procedure for the RL agent. Our approach leverages an existing controller for safe exploration in offline RL~\cite{offline_RL}, without requiring it to maximize our reward function.
        This is in contrast to a supervised learning approach like behavior cloning~\cite{travis_goal_conditioned, ross2011reduction}. \\
        In summary, our proposed approach involves the following steps:
        \begin{enumerate}
        \item First, we use PID controllers to explore the feasible action space. These controllers are employed with random setpoints to move the target to different locations.
        \item We use online RL with PID controllers in an outer region to prevent failures near the edges. This enhances stability and reduces the need to frequently rerun the trial, making the training process more efficient. Additionally, we frequently take random actions with a 0.3 probability when using the outer controller since we're confident it will stabilize the system. Lastly, we recommend starting with an outer region of $20\%$ and gradually decreasing it.
        \item Finally, the outer classic controller is removed, and the RL agent is used alone.
        \end{enumerate}

	The final module in the controller is the spiral search, which plays a crucial role in minimizing the need to rerun trials after failures. To implement this module, we follow the approach proposed in~\cite{ours}, which avoids random searching for the lost diver and instead starts searching from the most probable direction.
 
    \subsection{Simulation Environment}\label{simsec}
    For controlled repeatable assessment, we use a 
    commercial ROS2-based
    underwater simulator with basic hydrodynamics support.
    Fig. \ref{fig_sim_1} provides a view of the simulation environment. The simulator~\cite{IRsim}, called \textit{AquaSim},  can model the dynamics of
    the six-legged robot we employ and
    faithfully replicate the hardware control 
    stack and physical interactions across a
    range of different scenarios. This repeatable 
    environmental variability allows us to test and evaluate different methods of control, vision, navigation, etc. The simulator is based on the Unity engine and includes features such as hydrodynamics, sensor simulation, and visual realism.

    The simulator models underwater forces, including thrust from flippers' motion and hydrodynamic drag, using a model loosely based on \cite{giguere2006characterization}. Note, the hydrodynamics and physical modeling in \textit{AquaSim} is only an approximation to real-world conditions. To simulate linear drag, the following formula is utilized:
    \begin{equation}
    F_D = \frac{1}{2}\rho v^2 C_d A 
    \end{equation}
    where $F_D$ represents the drag force, $\rho$ is the density of the fluid, $v$ denotes the velocity of the robot, $C_d$ represents the drag coefficient, and $A$ represents the area. This is evaluated three times using the local-frame $XYZ$ axes to compute the drag force. 
    After calculating all three local-frame drag forces, they are converted back to the world-frame and applied to the rigid body. In the field, turbulence and
    vortex shedding are experienced and are impractical or
    impossible to fully model analytically, and only data driven approximation is used .

    In terms of buoyancy, \textit{AquaSim} utilizes a simplified approximation that is modeled as follows:
    \begin{equation}
    F_b = mass \times (-g) \times B_{Coef}
    \end{equation}
    \textit{AquaSim} adjusts the robot's buoyancy through a tuning parameter called $B_{Coef}$. This parameter can be modified to create positive or negative buoyancy as required. \textit{AquaSim} uses Unity's built-in drag calculations to implement angular drag. Furthermore, the simulator also models currents and wave motion.
    
	\section{Experimental Evaluation}
    \begin{table*}[]
     \centering
     \begin{tabularx}{\textwidth}{| c | c | c | c | c | c | c | }
        \hline
        \textbf{\textbf{History size}} & \textbf{5} & \textbf{10} & \textbf{20} &  \textbf{30} & \textbf{PID} \\
         \hline
         Yaw Expected Cumulative Reward  & 12937.08$\pm$9840.8   & 8514.11$\pm$3385.62 & \textbf{30596.37}$\pm$\textbf{9956.83} & 14823.06$\pm$8463.23 &  17978.57$\pm$13848.02  \\
        \hline
         Pitch Expected Cumulative Reward  & 21019.99$\pm$9747.28  & 8471.76$\pm$5295.07 & \textbf{30772.65}$\pm$\textbf{8706.96} & 11798.61$\pm$8063.64 & 13733.64$\pm$10840.88 \\
        \hline
         Tracking Length & 37634.33$\pm$14118.97  & 15658.14$\pm$7609.55 & \textbf{45151.0}$\pm$\textbf{8707.61} & 22701.0$\pm$12931.9 & 19361$\pm$14276 \\ 
        \hline
         Yaw Immediate Reward Average & 0.40 $\pm$ 0.28  & 0.57 $\pm$ 0.10 & \textbf{0.69}$\pm$\textbf{0.19} & 0.66$\pm$0.14 & \textbf{0.91}$\pm$\textbf{0.03} \\ 
        \hline
         Pitch Immediate Reward Average & 0.55 $\pm$ 0.08  & 0.50$\pm$0.09 & \textbf{0.68}$\pm$\textbf{0.12} & 0.49$\pm$0.16 & \textbf{0.69}$\pm$\textbf{0.05} \\  
        \hline
    \end{tabularx}   
    \caption{The table summarizes the results of our comparison study, where we tested different historical data lengths to highlight the importance of considering delay time. }
     \label{table_comparision}
 \end{table*}
    
We developed a ROS2-based pipeline for a real robot platform, testing it for the application of scuba diver tracking. Despite online optimization, the system's frequency remains consistent with our previous work~\cite{ours}. It's stable with a PID controller in shared control but challenging with pure RL due to unpredictable scenarios.

    Regarding real-world implementation, PID setup is simple but tuning is difficult. We created a list of gains, conducting a discrete search by running the robot with each gain set and evaluating metrics discussed later. Note, the RL agent training requires PID controller tuning. However, the RL agent is an adaptive controller, whereas PIDs may need repeated tuning for the same changes in the robot or the environment, causing instability.
    
    To enable comparison studies, we used  \textit{AquaSim} with a human character as the target. In this regard, we generated a dataset of 1200 images using augmentation methods and trained YOLOv7 on that.
    Our study involved three types of evaluations, as detailed below. \\
    - Comparing the RL agent's performance to separate PID controllers.\\
    - Evaluating the ability of the RL agent to adapt to changes in the robot's parameters. \\ 
    - Studying how adding detection confidence to the reward function enhances the RL agent's ability to improve the vision system by directing objects to more accurate detection locations.
    
    Numerous trials were carried out to ensure result repeatability and enhance reliability. Each trial had a predefined time limit for realism, concluding after tracking more than 45,000 frames (around 30 mins). 
    Furthermore, the reward function defined in equation (\ref{eq_reward}) is closely tied to the object's distance from the image plane's center. We use this as a metric to evaluate short-term controller performance. In contrast, we prioritize tracking length as the primary metric for long-term assessment. Additionally, we report the expected cumulative reward, which represents the average of discounted rewards collected during the trials. This provides insight into how well the controller balances transient and long-term performance.
    
    In our initial study, we did not consider the impact of detection confidence in our reward function, which led the RL agent to prioritize keeping the target centered. We also calculated the mean and standard deviation for the metrics mentioned earlier and evaluated performance separately for each direction. To create a challenging tracking task, the target was programmed to move randomly in all three directions at short-term fixed speeds, as illustrated in Fig. \ref{diver_behavior}. Despite the diver's complex movements in multiple directions, the RL agent, equipped with a history size of 20, demonstrated remarkable tracking performance. 

    \begin{figure}[]
        \centering
        \subfigure{\includegraphics[width=0.32\textwidth]{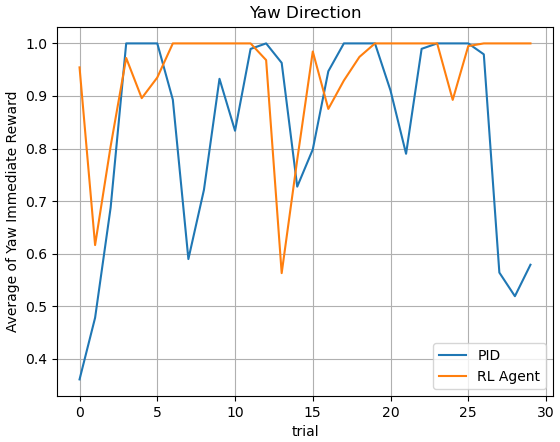}}\\
        \subfigure{\includegraphics[width=0.32\textwidth]{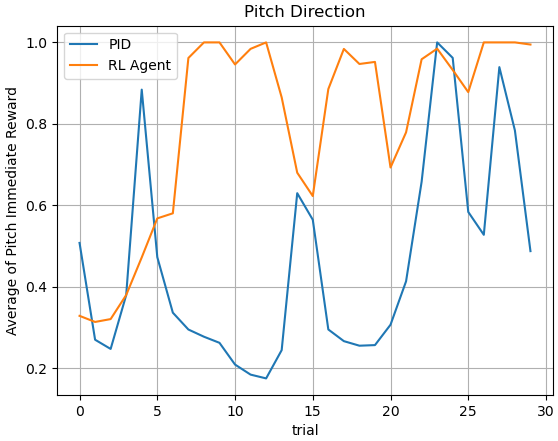}} 
        \caption{This outcome indicates that PID controllers struggle with handling changes in the robot's parameters as the error variance is high.  }
        \label{fig_study2}
    \end{figure}
    
    Table \ref{table_comparision} summarizes our comparison study, confirming our hypothesis that a history size lower than the delay time results in poor performance. With an appropriately chosen history size, the centralized RL agent consistently outperforms the separated PID controller over the long term, demonstrating superior performance in both pitch and yaw directions. The RL agent's ability to consider the history of target movements enhances the system's stability. However, sub-optimal policies can lead to unstable behavior due to the system being a higher-order Markov process. 
    The last two rows of the table emphasize that PIDs excel in transient behavior but face challenges in long-term tracking due to their independent operation, which can lead to conflicts. Additionally, increasing the history size may not necessarily improve performance; instead, it can slow down the controller and expand the state-action space for exploration.
    
    In our second study, we aim to assess the controllers' ability to handle model uncertainty and fault tolerance. The outcomes show that online improvements empower the RL agent to adapt to changes in the robot. We conducted two sets of trials: one involving increased angle damping and negative buoyancy in the robot, and another assuming a faulty leg. To delve into the behavior of the two controllers, we examined their transient performance by averaging immediate rewards across sequential trials. This comparison was feasible since the trial durations were identical. 
    These scenarios closely resemble real-world situations and mirror practical challenges. In both cases, the RL agent consistently outperformed the PIDs, demonstrating the effectiveness of our approach.
    
    Fig. \ref{fig_study2} shows that high-angle damping needs more aggressive controllers to compensate for errors. PIDs are insufficient and result in poor performance, with errors increasing. Negative buoyancy worsens this situation for pitch direction. However, the RL agent can improve online and generate required aggressive commands even with small errors, outperforming the PIDs and improving as it interacts with the environment, despite not being trained for such a situation. 
    
\begin{figure}[]
        \centering
        \subfigure{\includegraphics[width=0.32\textwidth]{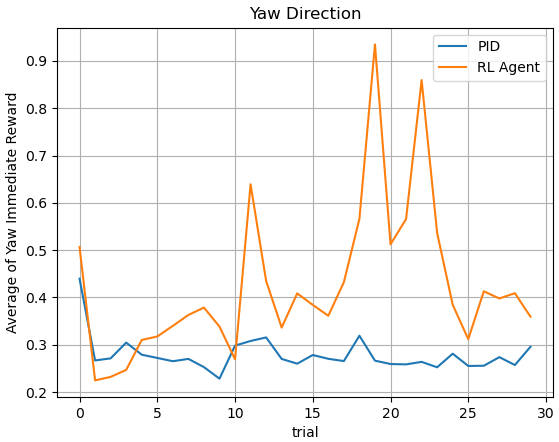}}\\
        \subfigure{\includegraphics[width=0.32\textwidth]{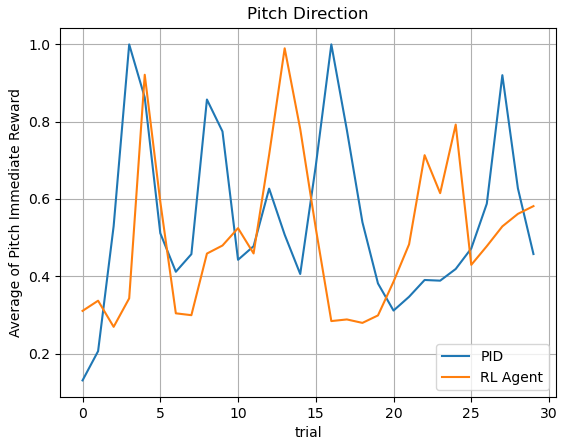}} 
        \caption{RL and PID responses to an induced
        right-rear leg actuator failure that impacts yaw. Note that RL yields superior rewards, and results are averaged across trials.}
        \label{fig_study2_2}
    \end{figure}

    For the second part of the study, we simulated a fault in the actuator of the right-rear leg of the robot, which led to a failure in receiving commands. Based on this, we conducted a trial to compare the performance of PIDs and the RL agent. The outcomes are summarized in Fig. \ref{fig_study2_2}. The RL agent gradually outperformed the PID controllers, particularly in the yaw direction, demonstrating its ability to adapt and improve in response to the situation.  

    \begin{figure}[]
        \centering        \subfigure{\includegraphics[width=0.3\textwidth]{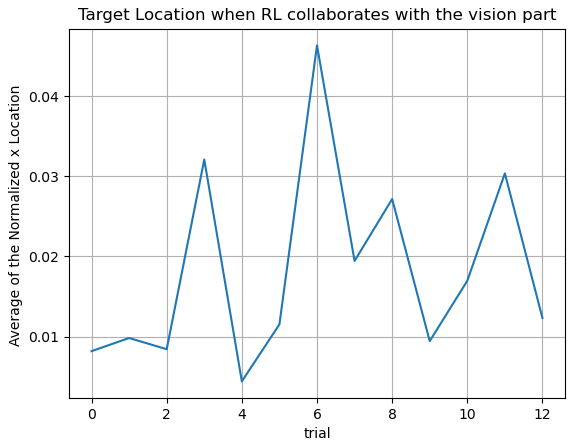}}\\
        \caption{The figure shows the RL agent's performance when the reward function considers detection confidence. In this case, the RL agent pushes the target rightward intentionally to help the vision module.}
        \label{fig_study3}
    \end{figure}
    Our final study demonstrates how an RL agent collaborates with a vision module, particularly useful when the vision model's training data is biased or influenced by factors like fog. We weakened the object detector's performance on the left side of the image plane, expecting the agent to move the target to the right side. By incorporating detection confidence into the reward function, we evaluated the RL agent's performance with history size of 20. The results in Fig. \ref{fig_study3} show the agent's ability to keep the target near zero or slightly positive in the x direction, improving detection accuracy and reducing failures.


    \section{Conclusion}
        We propose a hybrid tracking approach exemplified on a six-legged 6-DOF swimming robot. The paper explores the use of a centralized DQN-based  controller to generate yaw and pitch rates while maintainting a trade-off between transient and long-term tracking. The paper also introduces a  framework to train the agent while safely exploring the state-action space under classical control. Experimental results demonstrate that the RL agent outperforms  PID and successfully collaborates with the vision module. Additionally, the RL agent can adapt to changes in the model or environmental parameters, ensuring robust performance for underwater object tracking.

 
	{\small
		\bibliographystyle{ieeetr}
		\bibliography{egbib}
	}

\end{document}